\documentclass[sigconf]{acmart}

\usepackage{multirow}

\usepackage{makecell}    
\usepackage{colortbl}
\definecolor{light-gray}{gray}{0.92}  

\usepackage{subcaption}

\usepackage{marvosym}

\usepackage{tikz}
\usetikzlibrary{shapes.geometric}

\definecolor{realFill}{HTML}{55B83D}
\definecolor{realEdge}{HTML}{165B0F}

\definecolor{fakeFill}{HTML}{F4772B}
\definecolor{fakeEdge}{HTML}{7A2D0C}

\AtBeginDocument{
  }

\copyrightyear{2026}
\acmYear{2026}
\setcopyright{cc}
\setcctype{by}
\acmConference[MM '26]{Proceedings of the 34th ACM International Conference on Multimedia}{November 10--14, 2026}{Rio de Janeiro, Brazil}
\acmBooktitle{Proceedings of the 34th ACM International Conference on Multimedia (MM '26), November 10--14, 2026, Rio de Janeiro, Brazil}
\acmDOI{10.1145/3767308.3836365}
\acmISBN{979-8-4007-2213-4/2026/11}

\begin{document}

\title{Learning to Falsify: Hierarchical Multiple Instance Learning for High Resolution AI-Generated Image Detection}

\author{Haoran Chen}
\orcid{0009-0008-2249-444X}
\affiliation{%
  \institution{Zhejiang Gongshang University}
  \city{Hangzhou}
  \state{Zhejiang}
  \country{China}
}
\email{haoran@pop.zjgsu.edu.cn}

\author{Daiqi Gao}
\orcid{0009-0005-1276-5865}
\affiliation{%
  \institution{Zhejiang Gongshang University}
  \city{Hangzhou}
  \state{Zhejiang}
  \country{China}
}
\email{25020100007@pop.zjgsu.edu.cn}

\author{Xinyuan Li}
\orcid{0009-0004-1141-4759}
\affiliation{%
  \institution{Zhejiang Gongshang University}
  \city{Hangzhou}
  \state{Zhejiang}
  \country{China}
}
\email{2412190720@pop.zjgsu.edu.cn}

\author{Yongwei Wang}
\orcid{0000-0001-9712-8964}
\affiliation{%
  \institution{Zhejiang University}
  \city{Hangzhou}
  \state{Zhejiang}
  \country{China}
}
\email{yongwei.wang@zju.edu.cn}

\author{Lei Zhu}
\orcid{0000-0003-3871-663X}
\affiliation{%
  \institution{The Hong Kong University of Science and Technology (Guangzhou)}
  \city{Guangzhou}
  \state{Guangdong}
  \country{China}
}
\email{leizhu@hkust-gz.edu.cn}

\author{Rui Cai}
\orcid{0009-0006-2823-3051}
\affiliation{%
  \institution{Zhejiang Gongshang University}
  \city{Hangzhou}
  \state{Zhejiang}
  \country{China}
}

\email{cairuics@gmail.com}

\author{Jianfeng Dong}
\correspondingauthor

\orcid{0000-0001-5244-3274}

\affiliation{%
  \institution{Zhejiang Gongshang University}
  \city{Hangzhou}
  \state{Zhejiang}
  \country{China}
}
\affiliation{%
  \institution{Zhejiang Key Laboratory of Big Data and Future E-Commerce Technology}
  \city{Hangzhou}
  \state{Zhejiang}
  \country{China}
}
\email{dongjf24@gmail.com}
\thanks{\ding{41}\ Corresponding author}

\begin{abstract}

Recent advances in generative models have made synthetic images increasingly photorealistic, posing growing challenges to reliable and generalizable detection of AI-generated images. Existing methods typically rely on holistic image-level classification, which may overlook local artifacts, especially in high-resolution images. To address this limitation, we formulate this task as a falsification-driven multiple instance learning problem, where a single fake patch can falsify the whole image, while a real image requires all patches to be authentic. Based on this formulation, we propose \textit{Learning to Falsify (LTF)}, a hierarchical multiple instance learning framework that models an image as multiple bags of instances across different spatial levels and aggregates their predictions under the falsification-inspired principle. We further introduce Cross-Shuffle during training to encourage the model to learn local artifact evidence rather than image-level shortcuts. Furthermore, we propose the \textit{Falsification-driven Test-time Aggregation Strategy (FATAS)}, a plug-and-play module that extends the same falsification-inspired principle to existing detectors by aggregating patch-level predictions. Extensive experiments demonstrate that LTF achieves strong generalization and robustness across multiple benchmarks, while FATAS substantially improves representative detectors in high-resolution scenarios without retraining. Code is available at \url{https://github.com/HuiGuanLab/LTF.git}
\end{abstract}



\maketitle
\renewcommand{\shortauthors}{Haoran Chen et al.}

\begin{figure}[ht]
    \centering
    \includegraphics[width=\linewidth]{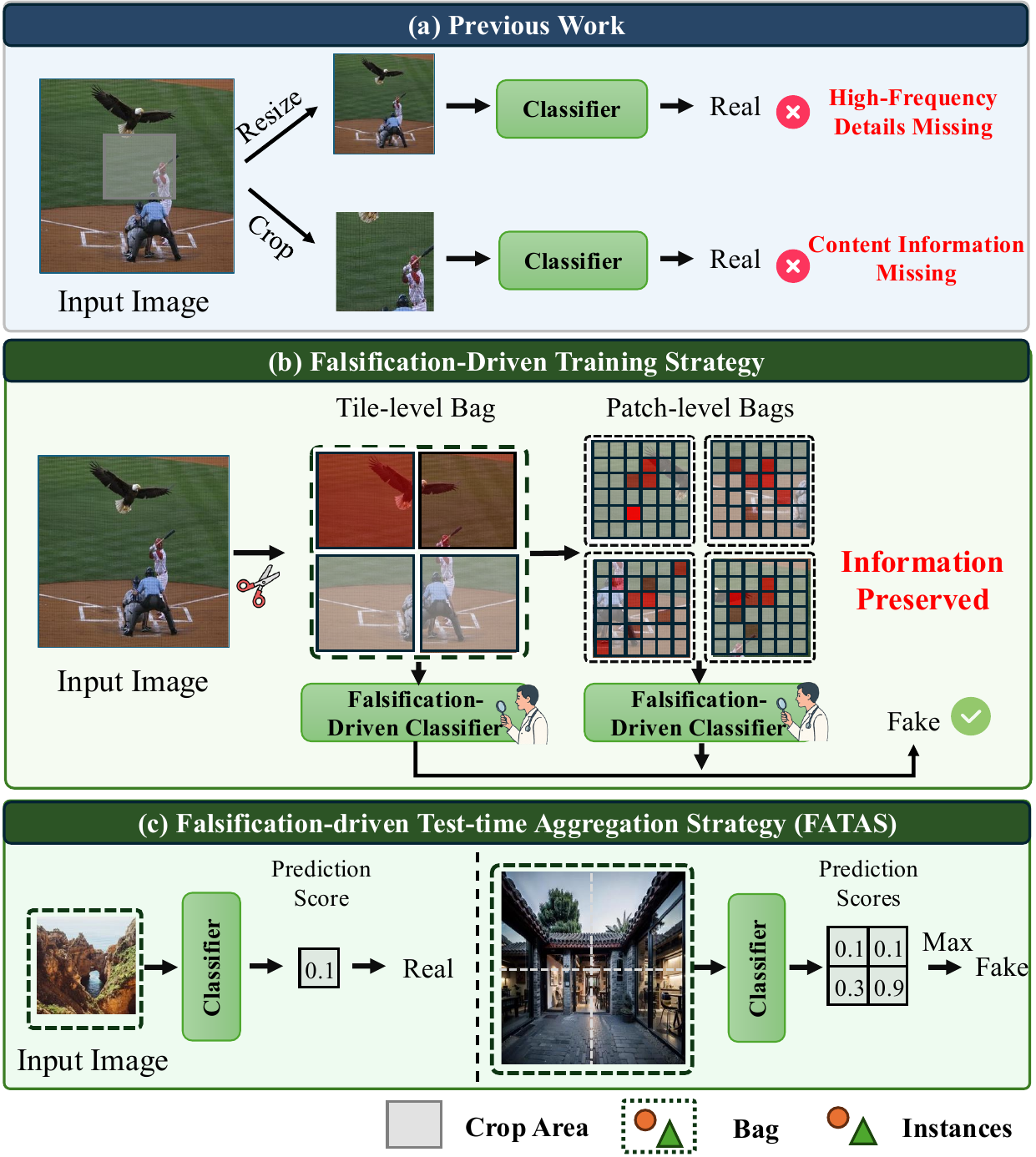}
    \caption{Motivation. (a) Conventional pipelines rely on image resizing or local cropping, which inevitably discard either fine-grained generated artifacts or content. (b) In contrast, we process the image at its original resolution and represent it as multiple bags of instances across multiple scales, thereby preserving both global context and fine-grained details. (c) During inference, image-level predictions are obtained via falsification-driven aggregation of instance-level scores, where a single decisive instance is sufficient to classify the image as fake. }
    \Description{A three-panel diagram contrasts conventional image-level detection with the proposed falsification-driven strategies. Panel a starts from a baseball photograph containing players and a bird. Resizing removes fine details, while a local crop removes contextual content; both altered inputs are nevertheless classified as real. Panel b keeps the original image, divides it into four tiles, further represents each tile as a grid of local instances, and sends the tile-level and patch-level bags to separate falsification-driven classifiers. A suspicious instance can make the final result fake while spatial information is preserved. Panel c contrasts single-crop inference with multi-patch inference. A single crop receives a score of 0.1 and is classified as real, whereas four local scores of 0.1, 0.1, 0.3, and 0.9 are combined by a maximum operation, allowing the high-scoring region to classify the image as fake.}
    \label{fig:motivation}
\end{figure}

\section{Introduction}
Recent advances in generative models, from GANs~\cite{goodfellow2014generative,radford2015unsupervised,brock2018large,karras2019style} to diffusion-based methods~\cite{ho2020denoising,dhariwal2021diffusion,nichol2021glide,rombach2022high}, have made it possible to synthesize highly photorealistic images at scale. As these synthetic images become increasingly indistinguishable from real photographs, their misuse poses growing societal risks, including privacy violations, identity forgery, and the spread of misinformation. Consequently, developing reliable and generalizable detectors for AI-generated images has become an urgent and important problem.

Despite substantial progress in AI-generated image detection~\cite{yan2024sanity,pmlr-v267-yan25b,li2025toward,li2026synerdetect,li2024unionformer}, most existing methods still follow a holistic image-level classification paradigm. To fit standard visual backbones~\cite{he2016deep,oquab2023dinov2}, they typically either resize the entire image to a fixed resolution~\cite{tan2024rethinking,wang2020cnn,frank2020leveraging} or use a single local crop as a proxy for the full image~\cite{chen2024single,pmlr-v267-yan25b,yuan2025mlep}. The resulting input is then compressed into a global representation for image-level classification. 
As illustrated in Fig.~\ref{fig:motivation}(a), this paradigm is poorly suited for detecting localized artifacts. Resizing may suppress fine-grained forensic cues, single-crop inference only provides partial spatial coverage, and global representations diminish the influence of artifacts confined to local regions. These limitations become more severe for high-resolution images, where fixed-size resizing causes substantial loss of local details and a single crop captures only a small fraction of the image.



To overcome this limitation, we draw inspiration from counterexample based falsification in mathematics, where a single counterexample suffices to refute a universal claim. We accordingly introduce the \textit{falsification}-inspired principle for AI-generated image detection: a single fake region suffices to reject image authenticity, whereas accepting an image as real requires all its regions to be authentic. 
Based on this principle, we propose \textit{Learning to Falsify (LTF)}, a hierarchical multiple instance learning framework for identifying localized generative artifacts. As illustrated in Fig.~\ref{fig:motivation}(b), LTF avoids resizing the image or using a single crop. Instead, it retains each image at its original resolution and partitions it without discarding pixels, thereby preserving full spatial coverage and fine-grained forensic details. These local regions are further organized into a hierarchical bag structure: each image forms a tile-level bag whose instances are image tiles, and each tile forms a patch-level bag composed of local patches. This design represents each image with one tile-level bag and multiple patch-level bags, enabling LTF to search for suspicious evidence at multiple granularities. Under the falsification-inspired principle, a few suspicious instances within any bag can provide sufficient evidence for a fake prediction, so the final image-level prediction is driven by localized artifact evidence rather than a holistic image representation.
However, since instances within the same bag are derived from the same image and share the same image-level label, the model may learn image-level shortcuts instead of truly identifying localized generative artifacts. To address this issue, we introduce Cross-Shuffle, which recomposes fake bags by mixing features from real images at both the tile and patch levels. This strategy breaks shortcut correlations among instances and forces the falsification-driven classifier to focus on genuine artifact evidence.

As shown in Fig.~\ref{fig:motivation}(c), FATAS first divides the input image into multiple tiles and predicts a fake score for each tile using an existing detector. It then aggregates tile-level predictions by emphasizing the most suspicious local regions. This design follows the falsification rule: once any tile exhibits strong generative artifacts, the whole image should be classified as fake, thereby preventing local fake evidence from being diluted by holistic modeling. Without retraining or architecture modification, FATAS substantially improves existing detectors, boosting the accuracy of CNNSpot~\cite{wang2020cnn} by over 20 percentage points on Midjourney and that of AIDE~\cite{yan2024sanity} by nearly 10 percentage points on Chameleon~\cite{yan2024sanity}.


Our contributions are summarized as follows:
\begin{itemize}
    \item We introduce the \textit{falsification-inspired} principle, under which a single fake region can falsify an image, whereas a real image requires all regions to be authentic. We instantiate this principle as \textit{LTF}, a hierarchical multiple-instance learning framework that explicitly learns localized discriminative artifacts. LTF incorporates Cross-Shuffle to create diverse instance combinations and improve model generalization.
    
    \item We introduce FATAS, a Falsification-driven Test-time Aggregation Strategy that requires no retraining and can be readily integrated into existing detectors at inference time. Despite its simplicity, it consistently improves detection performance in high-resolution scenarios.
    
    \item Extensive experiments on multiple cross-dataset benchmarks demonstrate that our method achieves state-of-the-art performance, with particularly strong gains on high-resolution datasets, validating both the effectiveness of the framework and its generalization capability.
\end{itemize}

\section{Related Work}
The increasing realism and diversity of AI-generated images have made it more difficult to develop detectors that are both reliable and generalizable across different synthesis techniques~\cite{wang2020cnn,wang2023dire,tan2024rethinking,shen2025beyond,shencore}. Existing research on AI-generated image (AIGI) detection is commonly divided into two main directions: \emph{low-level artifact modeling} and \emph{high-level semantic modeling}. 


\textit{Low-level approaches} focus on subtle statistical irregularities introduced by generative processes~\cite{wang2020cnn,tan2023learning,wang2023dire,chu2025fire,tan2024rethinking,yuan2025mlep,NEURIPS2024_6dddcff5}, based on the assumption that synthetic images contain traces absent from authentic photographs. For example, CNNSpot~\cite{wang2020cnn} improves cross-generator generalization through extensive data augmentation. LGrad~\cite{tan2023learning} exploits gradient-based cues, while DIRE~\cite{wang2023dire} measures diffusion reconstruction errors, a direction later extended by FIRE~\cite{chu2025fire} through frequency-domain guidance. NPR~\cite{tan2024rethinking} models abnormal pixel dependencies via local neighborhood analysis, MLEP~\cite{yuan2025mlep} extends this line of work with multi-granularity and entropy-based features, and SLC~\cite{jia2025secret} focuses on differences in color distributions. By learning transformed representations $\mathbf{f}(x)$, these methods extract robust forensic signatures beyond raw RGB values. Breaking~\cite{NEURIPS2024_6dddcff5} further shows that suppressing semantic information can improve cross-distribution generalization, underscoring the value of simple statistical cues.

\emph{High-level approaches} instead leverage pretrained visual representations to assess semantic plausibility and global visual consistency~\cite{ojha2023towards,tan2025c2p,pmlr-v235-chen24ay,zhang2025truemoe,li2024unionformer,li2026realnet,li2026synerdetect,shen2025beyond}. UnivFD~\cite{ojha2023towards}, for instance, adopts CLIP-ViT as its feature backbone to improve cross-generator robustness. Building on this paradigm, subsequent studies extend CLIP in different directions: FatFormer~\cite{liu2024forgery} augments it with frequency-aware adapters, C2P-CLIP~\cite{tan2025c2p} adapts it through image-text fine-tuning, and DRCT~\cite{pmlr-v235-chen24ay} improves global feature representations via hard-negative contrastive learning. Despite their differences in feature construction, both low-level and high-level methods still primarily derive image-level predictions from globally aggregated evidence, making it difficult to explicitly capture localized cues.

This issue becomes particularly acute in high-resolution regimes, where the cost of resizing and cropping is amplified: downscaling a high-resolution image to a fixed resolution compresses fine-grained details more aggressively, while a fixed-size crop covers an increasingly small portion of the scene. Existing detection methods for high-resolution AI-generated images (AIGI) generally follow one of three paradigms: resizing the entire image to a fixed resolution, analyzing a single local crop, or processing multiple patches followed by feature aggregation. Representative approaches in the first category include NPR~\cite{tan2024rethinking} and UnivFD~\cite{ojha2023towards}. DRCT~\cite{pmlr-v235-chen24ay} and SAFE~\cite{10.1145/3690624.3709392} instead adopt crop-based analysis, while more recent methods such as SPAI~\cite{karageorgiou2025any} and HiDA-Net~\cite{mu2025no} improve image coverage through multi-patch processing. Although these strategies partially mitigate the difficulty of handling high-resolution inputs, they primarily emphasize broader image coverage. The effective modeling and aggregation of local discriminative evidence remain insufficiently explored.

\begin{figure*}
    \centering
    \includegraphics[width=1\linewidth]{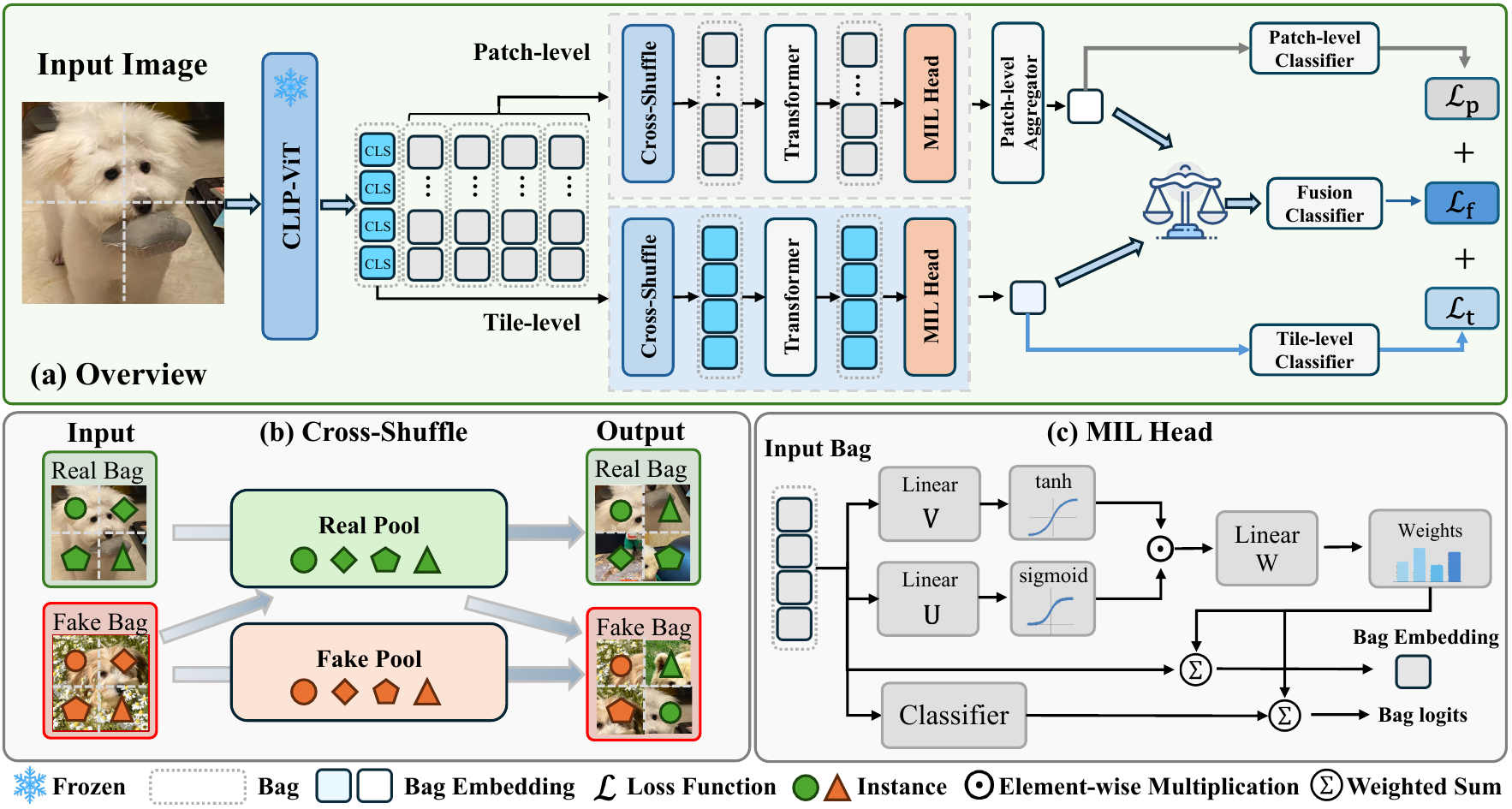}
    \caption{Overview of the proposed hierarchical falsification-driven detection framework. The input image is encoded into two levels of instances by a frozen CLIP-ViT. Cross-Shuffle enhances instance discrimination across bags, and the MIL head aggregates sparse local evidence at each level via gated attention pooling. The final prediction is obtained by fusing the two levels with multi-level supervision.}
    \Description{A three-part architecture diagram details the Learning to Falsify framework. In panel a, a frozen CLIP-ViT encodes a tiled input image into patch-level tokens and tile-level CLS tokens. Each level passes through its own Cross-Shuffle module, Transformer block, MIL head, and aggregator. Patch-level and tile-level representations feed their respective classifiers and are also combined by a learned fusion classifier. The three outputs are supervised by patch, tile, and fusion losses. Panel b illustrates Cross-Shuffle: instances from real and fake input bags are collected into separate real and fake pools, then sampled to form recomposed bags. A recomposed real bag contains only real instances, while a recomposed fake bag may mix real and fake instances but retains at least one fake instance. Panel c expands the gated-attention MIL head. Each input instance follows two branches, linear V followed by tanh and linear U followed by sigmoid; their element-wise product is mapped by linear W to attention weights. The weights aggregate instance features into a bag embedding and aggregate instance-classifier outputs into bag logits.}
    \label{fig:framework}
\end{figure*}

\section{Method}
As shown in Fig.~\ref{fig:framework}, we propose \textit{Learning to Falsify (LTF)}, a hierarchical detection framework designed to preserve full spatial coverage and identify localized artifacts that can falsify an image's authenticity. Given an input image, LTF first uses a frozen CLIP-ViT to construct hierarchical tile- and patch-level bags, enabling the model to represent both coarse regional context and fine-grained local details. To encourage the model to focus on local synthetic artifacts rather than image-level shortcuts, Cross-Shuffle recomposes instances across bags during training. The resulting bags are then processed by falsification-driven MIL heads, which is used to identify and aggregate discriminative local evidence. Finally, predictions from the tile and patch levels are fused and jointly optimized with multi-level supervision.


\subsection{Hierarchical Image Representation}
\label{sec:hierarchical_representation}


Instead of resizing the whole image or using only one crop, we keep the input image at its original resolution and represent it into hierarchical tile-level and patch-level features.
Specifically, given an input image, we first partition it into $N$ non-overlapping tiles of size $224 \times 224$, and then use a frozen CLIP-ViT backbone to encode each tile independently. For the $i$-th tile, the encoder outputs one CLS token $\mathbf{c}_i$ and $M$ patch tokens $\{\mathbf{p}_{i,j}\}_{j=1}^{M}$. The CLS token summarizes the overall content of the tile, while the patch tokens describe smaller local regions within the tile and retain fine-grained artifact cues.

Following the multiple instance learning (MIL) paradigm, we treat each token as an instance and organize these instances into hierarchical bags. 
For each image, all CLS tokens from the $N$ tiles are grouped into one tile-level bag, which captures coarse regional patterns. Meanwhile, the $M$ patch tokens within each tile form a separate patch-level bag, preserving fine-grained local patterns inside that tile.
Formally, these bags are defined as:
\begin{equation}
    \mathcal{B}^{\mathrm{tile}}
    =\{\mathbf{c}_i\}_{i=1}^{N},
    \qquad
    \mathcal{B}^{\mathrm{patch}}_i
    =\{\mathbf{p}_{i,j}\}_{j=1}^{M},
    \quad i=1,\ldots,N.
    \label{eq:hierarchical_bags}
\end{equation}
Thus, an image partitioned into \(N\) tiles is represented by one tile-level bag and \(N\) patch-level bags. In the example shown in Fig.~\ref{fig:framework}, \(N=4\), the image is represented as one tile-level bag and four patch-level bags. 

\subsection{Cross-Shuffle}
\label{sec:cross_shuffle}

Given the hierarchical bags constructed above, directly applying standard MIL optimization is still suboptimal. Since all instances in a bag inherit the same image-level label, the model may treat the bag as a globally labeled sample and memorize image-level semantic shortcuts, rather than learning which local instances contain discriminative generative artifacts. To overcome this, we propose Cross-Shuffle, which dynamically recomposes bags across images during training, as illustrated in Fig.~\ref{fig:framework}(b). The key idea is to create more diverse instance compositions while keeping the bag label strictly valid under the falsification principle: a real bag should only contain authentic instances, whereas a fake bag remains fake as long as at least one fake instance is preserved.

At each hierarchical level, we construct a real pool and a fake pool within each mini-batch by collecting instances from real and fake images, respectively. Given a replacement ratio $\rho\in[0,1)$, Cross-Shuffle randomly selects $\lfloor \rho K \rfloor$ positions in each bag for replacement. For a real bag, the selected instances are replaced only with instances from the real pool, ensuring that the recomposed bag remains fully authentic. For a fake bag, the selected instances are replaced with instances sampled from either the real or fake pool. Since $\rho<1$, at least one original instance from the source fake bag is preserved, so the recomposed fake bag remains label-consistent under the falsification principle. 
In this way, Cross-Shuffle breaks the shortcut correlations among instances from the same image while preserving valid bag labels. 

\subsection{Falsification-Driven MIL Aggregation}
\label{sec:mil_aggregation}


Given the recomposed bags from Cross-Shuffle, the remaining challenge is to infer the bag label from a set of instance features. Guided by the falsification-inspired principle, synthetic evidence in a fake bag is often highly localized, meaning uniform instance weighting easily dilutes these subtle cues among predominant authentic features. Therefore, an effective MIL aggregator must selectively identify decisive instances and magnify their contributions to the bag-level decision.

To this end, we introduce Falsification-Driven MIL Aggregation. As illustrated in Fig.~\ref{fig:framework}, we adopt a gated-attention MIL head~\cite{ilse2018attention} to process the contextualized instance features produced after Cross-Shuffle and Transformer encoding. The MIL head computes an importance score for each instance and normalizes these scores into attention weights. These weights are then used to aggregate both instance features and instance-level classification logits, producing a bag-level representation and bag-level logits, respectively. The resulting outputs are used for subsequent dual-branch fusion.


Specifically, given a recomposed bag $\widetilde{\mathcal{B}}$, a Transformer block first encodes its instances and produces a contextualized feature $\mathbf{h}_j\in\mathbb{R}^{d}$ for each instance $j\in\{1,\ldots,K\}$. The gated-attention module~\cite{ilse2018attention} then estimates the importance of each instance by computing an attention score $s_j$ as follows:
\begin{equation}
    s_j
    =
    \mathbf{w}^{\top}
    \left[
        \tanh(\mathbf{V}\mathbf{h}_j)
        \odot
        \sigma(\mathbf{U}\mathbf{h}_j)
    \right],
    \label{eq:attention_score}
\end{equation}
where $\tanh(\cdot)$ and $\sigma(\cdot)$ denote element-wise hyperbolic tangent and sigmoid functions, respectively, $\odot$ denotes element-wise multiplication, and $\mathbf{V},\mathbf{U}\in\mathbb{R}^{L\times d}$, $\mathbf{w}\in\mathbb{R}^{L}$ are learnable parameters with hidden dimension $L$.
The attention scores are normalized across all $K$ instances using
softmax:
\begin{equation}
    a_j
    =
    \frac{\exp(s_j)}
    {\sum_{k=1}^{K}\exp(s_k)},
    \qquad
    j\in\{1,\ldots,K\},
    \label{eq:attention_weight}
\end{equation}
The resulting weight $a_j$ measures the relative contribution of the $j$-th instance to the bag-level prediction. Therefore, instances assigned higher attention scores contribute more strongly during aggregation. Under the falsification-inspired principle, this weighting mechanism allows localized synthetic evidence to dominate the bag-level decision.




While the attention weights identify which instances are important, the MIL head also needs to determine what each important instance supports, \textit{i.e.}, real or fake. Therefore, we attach a lightweight linear classifier to each contextualized instance feature $\mathbf{h}_j$ to produce instance-level binary classification logits $\boldsymbol{\ell}_j$. These logits provide explicit local authenticity evidence, while the attention weights determine how strongly each instance should contribute to the bag-level decision. The bag-level embedding $\mathbf{h}_{\mathrm{bag}}$ and bag-level logits $\boldsymbol{\ell}_{\mathrm{bag}}$ are then computed by weighted aggregation:
\begin{equation}
    \mathbf{h}_{\mathrm{bag}}
    =
    \sum_{j=1}^{K}a_j\mathbf{h}_j,
    \qquad
    \boldsymbol{\ell}_{\mathrm{bag}}
    =
    \sum_{j=1}^{K}a_j\boldsymbol{\ell}_j.
    \label{eq:mil_aggregation}
\end{equation}

The bag-level logits produced at each hierarchical level are supervised by the image-level label using cross-entropy loss, and the resulting losses are summed as $\mathcal{L}_{\mathrm{bag}}$. This objective encourages the instance classifier and attention weights to jointly discover discriminative local artifacts, while providing reliable intermediate prediction signals for final prediction.

\begin{table*}[ht]
\centering
\caption{Performance on the GenImage Dataset. All methods are trained on the SDv1.4 subset of the GenImage. The best result in each accuracy column is highlighted in bold, and the second-best is underlined. For standard deviation (std), lower is better.}

\resizebox{1.0\textwidth}{!}{
    \begin{tabular}{l c c c c c c c c c | c c}
    \toprule
        Method & Venue & Midjourney & SDv1.4 & SDv1.5 & ADM & GLIDE & Wukong & VQDM & BigGAN & mAcc & std\\
    \midrule
ResNet-50~\cite{he2016deep}  &  CVPR2016   & 54.9 & 99.9 & 99.7 & 53.5 & 61.9 & 98.2 & 56.6 & 52.0 & 72.1 & 22.6 \\
DeiT-S~\cite{touvron2021training}& ICML2021 & 55.6 & 99.9 & 99.8 & 49.8 & 58.1 & 98.9 & 56.9 & 53.5 & 71.6 & 23.2\\
Swin-T~\cite{liu2021swin}    &  ICCV2021   & 62.1 & 99.9 & 99.8 & 49.8 & 67.6 & 99.1 & 62.3 & 57.6 & 74.8 & 21.1\\
CNNSpot~\cite{wang2020cnn}    &  CVPR2020  & 52.8 & 96.3 & 95.9 & 50.1 & 39.8 & 78.6 & 53.4 & 46.8 & 64.2 & 22.6\\
Spec~\cite{zhang2019detecting} & WIFS2019  & 52.0 & 99.4 & 99.2 & 49.7 & 49.8 & 94.8 & 55.6 & 49.8 & 68.8 & 24.1\\
F3Net~\cite{qian2020thinking}  & ECCV2020  & 50.1 & 99.9 & 99.9 & 49.9 & 50.0 & 99.9 & 49.9 & 49.9 & 68.7 & 25.8\\
GramNet~\cite{liu2020global}   & CVPR2020  & 54.2 & 99.2 & 99.1 & 50.3 & 54.6 & 98.9 & 50.8 & 51.7 & 69.9 & 24.2\\
UnivFD~\cite{ojha2023towards}  & CVPR2023  & 93.9 & 96.4 & 96.2 & 71.9 & 85.4 & 94.3 & 81.6 & 90.5 & 88.8 & 8.6\\
NPR~\cite{tan2024rethinking}  &  CVPR2024 & 81.0 & 98.2 & 97.9 & 76.9 & 89.8 & 96.9 & 84.1 & 84.2 & 88.6 & 8.3\\
FreqNet~\cite{tan2024frequency} & AAAI2024 & 89.6 & 98.8 & 98.6 & 66.8 & 86.5 & 97.3 & 75.8 & 81.4 & 86.8 & 11.6\\
FatFormer~\cite{liu2024forgery} & CVPR2024 & 92.7 & \textbf{100.0} & \textbf{99.9} & 75.9 & 88.0 & \textbf{99.9} & \underline{98.8} & 55.8 & 88.9 & 15.7\\
DRCT~\cite{pmlr-v235-chen24ay} & ICML2024 & 91.5 & 95.0 & 94.4 & 79.4 & 89.1 & 94.6 & 90.0 & 81.6 & 89.4 & 5.9\\
Breaking~\cite{NEURIPS2024_6dddcff5} & NeurIPS2024 & 83.9 & 98.9 & 93.0 & \textbf{99.1} & 97.7 & 85.4 & 92.7 & 90.5 & 92.7 & 5.8\\
AIDE~\cite{yan2024sanity} &ICLR2025 &79.4 &99.7 &99.8 &78.5 & 91.8 &98.7 &80.3& 66.9 &86.9 &12.3\\
VIB~\cite{11092328} & CVPR2025 & 88.1 & 99.6 & 99.6 & 73.9 & 74.3 &98.3 &89.4  & 91.2 & 89.3 &10.4   \\
Beyond~\cite{11095183} & CVPR2025 & 95.0 & 95.3 & 95.5 & 95.3 & 95.2 & 95.6 & 95.1 & 95.2 & 95.3 & \textbf{0.2} \\
Effort~\cite{pmlr-v267-yan25b} & ICML2025 & 82.4 & 99.8 & 99.8 & 78.7 & 93.3 & 97.4 & 91.7 & 77.6 & 90.1 & 11.8\\
SAFE~\cite{10.1145/3690624.3709392} & KDD2025 & \underline{95.3} & 99.4 & 99.3 & 82.1 & 96.3 & 98.2 & 96.3 & 97.8 & 95.6 & 5.6\\
C2P-CLIP~\cite{tan2025c2p} & AAAI2025 & 88.2 & 90.9 & 97.9 & 96.4 & \textbf{99.0} & 98.8 & 96.5 & \textbf{98.7} & 95.8 & 4.0\\
HSIC~\cite{yang2026hsic} &ICLR2026 & 80.8 & \underline{99.9} & \underline{99.8} & 91.1 & 97.0 & \underline{99.8} & \textbf{99.4} & 91.7 & 95.0 & 6.8\\
PPL~\cite{yang2026all} &ICLR2026 & 94.8 & 98.5 & 98.3 & 94.7 & 96.1 & 98.6 & 98.5 & 98.0 & \underline{97.2} & 1.7\\

\hline

\rowcolor{light-gray} LTF & This work & \textbf{97.3} & 98.9 & 98.6 & \underline{98.4} & 98.1 & 98.9 & 98.7 & \underline{98.1}  & \textbf{98.4} & \underline{0.5} \\

\bottomrule
    \end{tabular}
  }
\label{tab:SOTA1}
\end{table*}

\begin{table*}[ht]
\centering
\caption{Performance on the DRCT dataset.
All methods are trained on the SDv1.4 subset of DRCT. Results of
previous methods are obtained from DRCT~\cite{pmlr-v235-chen24ay}.}
\label{tab:SOTA2}

\resizebox{\textwidth}{!}{%
\begin{tabular}{lcccccccccccccccc|cc}
\toprule
\multirow{2}{*}{\textbf{Method}}
& \multicolumn{6}{c}{\textbf{SD Variants}}
& \multicolumn{2}{c}{\textbf{Turbo Variants}}
& \multicolumn{2}{c}{\textbf{LCM Variants}}
& \multicolumn{3}{c}{\textbf{ControlNet Variants}}
& \multicolumn{3}{c|}{\textbf{DR Variants}}
& \multirow{2}{*}{\textbf{mAcc}}
& \multirow{2}{*}{\textbf{std}} \\
\cmidrule(lr){2-7}
\cmidrule(lr){8-9}
\cmidrule(lr){10-11}
\cmidrule(lr){12-14}
\cmidrule(lr){15-17}

& \thead{LDM}
& \thead{SDv1.4}
& \thead{SDv1.5}
& \thead{SDv2}
& \thead{SDXL}
& \thead{SDXL-\\Refiner}
& \thead{SD-\\Turbo}
& \thead{SDXL-\\Turbo}
& \thead{LCM-\\SDv1.5}
& \thead{LCM-\\SDXL}
& \thead{SDv1-\\Ctrl}
& \thead{SDv2-\\Ctrl}
& \thead{SDXL-\\Ctrl}
& \thead{SDv1-\\DR}
& \thead{SDv2-\\DR}
& \thead{SDXL-\\DR}
& & \\
\midrule

CNNSpot~\cite{wang2020cnn}
& \underline{99.87}
& 99.91
& 99.90
& \underline{97.55}
& 66.25
& 86.55
& 86.15
& 72.42
& 98.26
& 61.72
& 97.96
& 85.89
& 82.84
& 60.93
& 51.41
& 50.28
& 81.12
& 17.6 \\

F3Net~\cite{qian2020thinking}
& 99.85
& 99.78
& 99.79
& 88.66
& 55.85
& 87.37
& 68.29
& 63.66
& 97.39
& 54.98
& 97.98
& 72.39
& 81.99
& 65.42
& 50.39
& 50.27
& 77.13
& 18.1 \\

CLIP/RN50~\cite{radford2021learning}
& 99.00
& \underline{99.99}
& \underline{99.96}
& 94.61
& 62.08
& 91.43
& 83.57
& 64.40
& 98.97
& 57.43
& \underline{99.74}
& 80.69
& 82.03
& 65.83
& 50.67
& 50.47
& 80.05
& 18.3 \\

GramNet~\cite{liu2020global}
& 99.40
& 99.01
& 98.84
& 95.30
& 62.63
& 80.68
& 71.19
& 69.32
& 93.05
& 57.02
& 89.97
& 75.55
& 82.68
& 51.23
& 50.01
& 50.08
& 76.62
& 17.0 \\

De-fake~\cite{sha2023fake}
& 92.10
& 99.53
& 99.51
& 89.65
& 64.02
& 69.24
& 92.00
& 93.93
& 99.13
& 70.89
& 58.98
& 62.34
& 66.66
& 50.12
& 50.16
& 50.00
& 75.52
& 18.4 \\

Conv-B~\cite{liu2022convnet}
& \textbf{99.97}
& \textbf{100.0}
& \textbf{99.97}
& 95.84
& 64.44
& 82.00
& 80.82
& 60.75
& \underline{99.27}
& 62.33
& \textbf{99.80}
& 83.40
& 73.28
& 61.65
& 51.79
& 50.41
& 79.11
& 18.3 \\

UnivFD~\cite{ojha2023towards}
& 98.30
& 96.22
& 96.33
& 93.83
& 91.01
& 93.91
& 86.38
& 85.92
& 90.44
& 88.99
& 90.41
& 81.06
& 89.06
& 51.96
& 51.03
& 50.46
& 83.46
& 17.0 \\

DRCT~\cite{pmlr-v235-chen24ay}
& 94.45
& 94.35
& 94.24
& 95.05
& 95.61
& 95.38
& 94.81
& 94.48
& 91.66
& 95.54
& 93.86
& 93.48
& 93.54
& 84.34
& \underline{83.20}
& \underline{67.61}
& \underline{91.35}
& \underline{7.11} \\

Beyond~\cite{11095183}
& 97.13
& 97.13
& 97.13
& 97.13
& \underline{97.13}
& \underline{97.13}
& \underline{97.13}
& \underline{97.13}
& 97.13
& \underline{97.13}
& 97.08
& \underline{96.81}
& \underline{94.73}
& \underline{92.26}
& 51.79
& 49.73
& 90.86
& 15.7 \\

\midrule

\textbf{LTF}
& 99.68
& 99.70
& 99.70
& \textbf{99.70}
& \textbf{99.70}
& \textbf{99.65}
& \textbf{99.65}
& \textbf{99.68}
& \textbf{99.69}
& \textbf{99.70}
& 99.69
& \textbf{99.68}
& \textbf{99.67}
& \textbf{99.70}
& \textbf{98.00}
& \textbf{82.80}
& \textbf{98.62}
& \textbf{4.08} \\

\bottomrule
\end{tabular}%
}
\end{table*}

\subsection{Hierarchical Classifiers}

The MIL heads produce hierarchical artifact evidence at two levels. At the tile level, the MIL head aggregates all tile instances into a tile-level image representation $\mathbf{h}_{\mathrm{tile}}$. At the patch level, the MIL heads produce $N$ patch-bag representations, each corresponding to one tile. We additionally introduce a patch-level aggregator with a learnable query token to summarize these $N$ representations into a unified patch-level image representation $\mathbf{h}_{\mathrm{patch}}$.

After obtaining the tile-level and patch-level image representations, we attach a classifier to each branch to make both levels directly discriminative. The tile-level classifier operates on $\mathbf{h}_{\mathrm{tile}}$ and captures coarse regional artifact evidence, while the patch-level classifier operates on $\mathbf{h}_{\mathrm{patch}}$ and focuses on fine-grained local artifacts. 

Since the reliability of tile-level and patch-level evidence may vary across images, we further adopt an adaptive fusion strategy. Specifically, a lightweight gating network takes the two branch logits as input and predicts a sample-adaptive fusion weight $\alpha\in(0,1)$, which indicates the relative contribution of the patch-level branch. The fused representation is then computed as
\begin{equation}
\mathbf{h}_{\mathrm{fuse}}
=
\alpha\mathbf{h}_{\mathrm{patch}}
+
(1-\alpha)\mathbf{h}_{\mathrm{tile}}.
\label{eq:tile_patch_fusion}
\end{equation}
Finally, $\mathbf{h}_{\mathrm{fuse}}$ is fed into a linear classifier to produce the image-level real/fake prediction.

The three classifiers are supervised by the image-level label, yielding the patch-level loss $\mathcal{L}_{p}$, tile-level loss $\mathcal{L}_{t}$, and fusion loss $\mathcal{L}_{f}$. The overall hierarchical classification objective is
\begin{equation}
\mathcal{L}_{\mathrm{cls}}
=
\mathcal{L}_{p}
+
\mathcal{L}_{t}
+
\mathcal{L}_{f}.
\label{eq:classification_objective}
\end{equation}
All classification
objectives are implemented using binary cross-entropy loss.

\subsection{Model Training and Inference}
\label{sec:training}


The model is optimized with two complementary objectives. The bag-level loss $\mathcal{L}_{\mathrm{bag}}$ directly supervises the intermediate MIL logits to strengthen localized artifact learning while the hierarchical classification loss $\mathcal{L}_{\mathrm{cls}}$ supervises the patch-level, tile-level, and fused image-level predictions defined above. The total training objective is defined as
\begin{equation}
\mathcal{L}_{\mathrm{total}}
=
\mathcal{L}_{\mathrm{bag}}
+
\mathcal{L}_{\mathrm{cls}},
\label{eq:training_objective}
\end{equation}
where the two losses are assigned equal weights.

After the model is trained, the output of the fusion classifier is used as the final real/fake prediction. It is worth noting that Cross-Shuffle is disabled during inference.

\subsection{Falsification-Driven Test-Time Aggregation Strategy}
\label{sec:fatas}

Building on the falsification-driven principle, we further explore whether existing detectors can be adapted to high-resolution AIGI detection at test time without retraining. Standard resizing or single-crop inference may suppress or omit local artifacts in high-resolution images. To address this issue, we propose a \textit{Falsification-Driven Test-Time Aggregation Strategy (FATAS)}, which preserves full spatial coverage and allows highly suspicious local regions to directly influence the image-level prediction. FATAS is plug-and-play and can be applied to off-the-shelf detectors without architectural modifications or re-training.

Specifically, given a high-resolution image, FATAS partitions it into $N$ non-overlapping tiles with full spatial coverage. An off-the-shelf detector then evaluates each tile $T_i$ independently and predicts a synthetic probability $p_i$. Instead of averaging all tile scores, which may dilute fake evidence, FATAS follows the falsification principle and aggregates the most suspicious tiles. The final image-level prediction is computed by averaging the top-$k$ tile probabilities:
\begin{equation}
\widehat{p}_{\mathrm{img}}
=
\operatorname{Mean}
\left(
\operatorname{Top}\text{-}k
(p_1,\dots,p_N)
\right),
\label{eq:fatas}
\end{equation}
where $\operatorname{Top}\text{-}k(\cdot)$ selects the $k$ highest probabilities among all $N$ tiles, $\operatorname{Mean}(\cdot)$ denotes the average operation.

By focusing on the most suspicious local regions while retaining full image coverage, FATAS reduces the risk of missing local artifacts in high-resolution inputs. It therefore extends the falsification-driven principle to test-time aggregation and provides a training-free strategy for improving existing detectors in high-resolution scenarios.


    
    
    


\section{Experiments}
\subsection{Setting}
\paragraph{Datasets.}
We conduct experiments on four benchmarks: GenImage~\cite{NEURIPS2023_f4d4a021}, DRCT~\cite{pmlr-v235-chen24ay}, Chameleon~\cite{yan2024sanity}, and HiRes-50K~\cite{mu2025no}. GenImage and DRCT are widely used benchmarks for evaluating the cross-dataset generalization ability of AIGI detectors, containing 8 and 16 test subsets, respectively. Chameleon provides a more realistic and challenging evaluation setting for high-resolution AIGI detection, comprising 26,033 samples with resolutions ranging from 0.5K to 4K. HiRes-50K is a dedicated benchmark for high-resolution AI-generated image detection across different resolution ranges. We compare our method with prior approaches on all four benchmarks. 

\paragraph{Metrics.}
We adopt Accuracy (Acc) and Average Precision (AP) for evaluation, consistent with prior studies~\cite{NEURIPS2023_f4d4a021,pmlr-v235-chen24ay,yang2026all}. A fixed threshold of 0.5 is used for binary classification decisions. For conciseness, we report Acc in the main paper and provide the corresponding AP results in the supplementary material.

\paragraph{Implementation Details.}
All experiments are conducted using a pretrained CLIP ViT-L/14 visual backbone, whose parameters remain frozen throughout training. The model is trained for 20 epochs using AdamW with a weight decay of 0.05 and a per-GPU batch size of 64. The main experiments are conducted on four NVIDIA A100 GPUs, each with 80\,GB of memory, resulting in a global batch size of 256. The learning rate is set to $4\times10^{-4}$ with a constant schedule. For data augmentation, we apply only JPEG compression and Gaussian blur. Additional implementation settings are provided in the supplementary material.

\begin{table*}[t]
  \centering
    \caption{Performance on the HiRes-50K Dataset across different resolutions. $^*$ denotes results reproduced by us using the official code under the same experimental settings~\cite{mu2025no}.
    }
  \label{tab:hires50k}
\setlength{\tabcolsep}{3pt}
\renewcommand{\arraystretch}{1.05}

\resizebox{\textwidth}{!}{%
\begin{tabular}{l|cccccccc|c}
\toprule
\textbf{Resolution Range}
& \textbf{0--900}
& \textbf{900--1200}
& \textbf{1200--1500}
& \textbf{1500--2000}
& \textbf{2000--2500}
& \textbf{2500--3000}
& \textbf{3000--5000}
& \textbf{$[5000,\infty)$}
& \textbf{Avg.} \\
\midrule

CNNSpot
& 58.46
& 56.37
& 58.73
& 63.14
& 67.42
& 60.90
& 66.30
& 58.90
& 61.28 \\

FreDetect
& 63.67
& 55.63
& 57.31
& 58.90
& 58.16
& 67.78
& 57.23
& 51.06
& 58.72 \\

UnivFD
& 67.00
& 58.35
& 62.20
& 65.95
& 66.05
& 58.20
& 58.15
& 60.05
& 61.99 \\

DIRE
& 59.11
& 64.25
& 62.10
& 66.66
& 75.84
& 63.40
& 69.31
& 62.18
& 65.36 \\

AIDE
& 65.87
& 57.29
& 49.90
& 58.23
& 51.88
& 65.31
& 61.04
& 42.16
& 56.46 \\

DRCT(ConvB)
& 65.30
& 66.19
& 68.78
& \underline{68.85}
& 68.78
& \underline{75.79}
& 69.65
& 54.03
& 67.17 \\

TextureCrop(CNNDetect)
& 57.45
& 52.81
& 59.83
& 63.17
& 65.31
& 55.56
& \underline{70.65}
& 69.49
& 61.78 \\

HiDA$^\ast$
& \underline{73.05}
& \underline{69.20}
& \underline{70.21}
& 68.82
& \underline{76.73}
& 62.00
& 58.40
& \underline{69.84}
& \underline{68.53} \\

\midrule

\textbf{LTF}
& \textbf{75.89}
& \textbf{76.55}
& \textbf{74.30}
& \textbf{86.15}
& \textbf{91.69}
& \textbf{86.60}
& \textbf{81.90}
& \textbf{78.39}
& \textbf{81.43} \\

\bottomrule
\end{tabular}%
}
\end{table*}

\subsection{Comparison with State-of-the-Art Methods}
\noindent\textbf{Generalization on GenImage and DRCT.}
To comprehensively assess the generalization of LTF across diverse generative models and data distributions, we conduct evaluations on the GenImage and DRCT benchmarks.
Tab.~\ref{tab:SOTA1} summarizes the performance comparison on GenImage, where all methods are trained on SD v1.4 subset. Our proposed LTF achieves the best performance, outperforming recent strong baselines such as PPL (97.2\%), C2P-CLIP (95.8\%), and SAFE (95.6\%). Moreover, LTF achieves the second-lowest standard deviation while substantially outperforming the method with the lowest standard deviation, demonstrating a better balance between performance and stability. Notably, the Midjourney subset contains the highest-resolution images in the GenImage benchmark, and LTF achieves the best accuracy on this subset as well, reaching 97.3\%, which further reflects its advantage in handling higher-resolution generated images.

Additionally, we conduct an experiment on DRCT, 
where all methods are trained on the SD v1.4 subset of DRCT-2M~\cite{pmlr-v235-chen24ay}. As shown in Tab.~\ref{tab:SOTA2}, LTF achieves the highest mean accuracy of 98.62\%, outperforming DRCT (91.35\%) and Beyond (90.86\%). These results demonstrate that LTF generalizes effectively across different data distributions.

\begin{figure}
    \centering
    \resizebox{0.38\textwidth}{!}{
    \includegraphics[width=\linewidth]{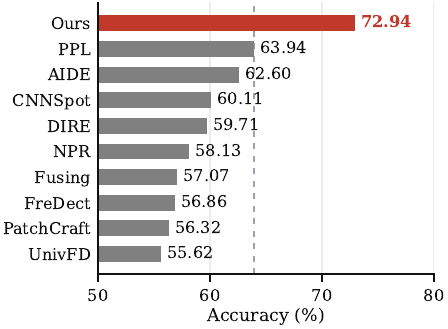}
    }
    \caption{Performance on high-resolution Chameleon benchmark, where all models are trained on SD v1.4 of GenImage.}
    \Description{A horizontal bar chart ranks ten detectors by accuracy on Chameleon. The proposed method, shown in red, ranks first at 72.94 percent. The nine gray baseline bars are PPL at 63.94, AIDE at 62.60, CNNSpot at 60.11, DIRE at 59.71, NPR at 58.13, Fusing at 57.07, FreDetect at 56.86, PatchCraft at 56.32, and UnivFD at 55.62 percent. The proposed method leads the strongest baseline, PPL, by 9.00 percentage points.}
    \label{fig:chameleon}
\end{figure}

\noindent\textbf{Performance on high-resolution datasets.}
In this experiment, we investigate the effectiveness of LTF in detecting high-resolution generated images.
We first conduct an evaluation on the Chameleon dataset. As shown in 
Fig.~\ref{fig:chameleon}, LTF achieves an accuracy of 72.94\%, outperforming the second-best method by a substantial margin. 
We further evaluate on HiRes-50K~\cite{mu2025no} to examine whether its advantage remains consistent under different image resolutions. As reported in Tab.~\ref{tab:hires50k}, LTF achieves the highest average accuracy of 81.43\% and ranks first across all resolution ranges. More importantly, its performance advantage becomes more pronounced on higher-resolution images, especially in resolution ranges spanning 1,500–5,000 pixels. Overall, these results demonstrate the effectiveness of LTF for detecting high-resolution AI-generated images.

\subsection{Experimental Evaluation of FATAS}
We further evaluate the effectiveness of our \emph{Falsification-Driven Test-Time Aggregation Strategy (FATAS)} on two high-resolution benchmarks, \emph{Chameleon} and \emph{Midjourney}. FATAS is applied in a training-free manner to five representative detectors, including NPR~\cite{tan2024rethinking}, AIDE~\cite{yan2024sanity}, CNNSpot~\cite{wang2020cnn}, DIRE~\cite{wang2023dire}, and Effort~\cite{pmlr-v267-yan25b}, using their official pre-trained checkpoints. 

\begin{figure}[t]
    \centering
    \includegraphics[width=\linewidth]{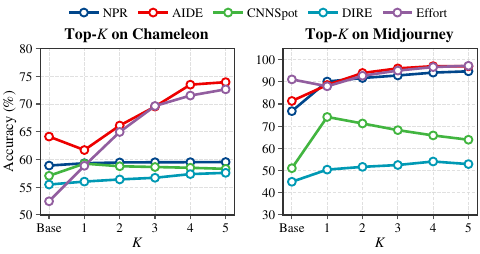}
    \caption{Performance of FATAS across inference settings. ``Base'' denotes the vanilla detector.}
    
    \Description{Two line charts plot accuracy against the number K of selected patches for NPR, AIDE, CNNSpot, DIRE, and Effort, with Base denoting each detector's original prediction. On Chameleon, AIDE and Effort improve most as K increases: AIDE rises from about 64 percent at Base to about 74 percent at K equals 5, and Effort rises from about 53 to about 73 percent. NPR, CNNSpot, and DIRE remain near 56 to 59 percent with smaller changes. On Midjourney, every detector is above its Base result for K of at least 2. NPR, AIDE, and Effort converge to the mid-to-high 90 percent range by K equals 5. CNNSpot jumps from about 51 percent at Base to about 74 percent at K equals 1 and then gradually declines to about 64 percent at K equals 5, still above Base. DIRE increases from about 45 percent at Base to the low 50 percent range. Most curves flatten after approximately three patches.}
\label{fig:tta_acc}
\end{figure}
Fig.~\ref{fig:tta_acc} shows that FATAS improves the majority of the evaluated detectors over their base results on both datasets. The gains are particularly significant on \emph{Midjourney}, where all five detectors achieve higher accuracy when $K\geq2$. On \emph{Chameleon}, substantial improvements are observed for AIDE and Effort, while CNNSpot also benefits consistently from multi-patch inference. Notably, these gains are achieved across detectors with substantially different base accuracies, demonstrating the broad applicability of FATAS. Performance generally becomes stable as more patches are included, indicating that a small number of informative patches is sufficient to improve prediction. A more comprehensive evaluation and detailed analysis are provided in the supplementary material. 

\subsection{Ablation Study}
We conduct two ablation studies on GenImage to evaluate the key designs
of LTF. Unless otherwise specified, models are trained on the SD v1.4
subset and evaluated on the full test set.

\noindent\textbf{Effect of Cross-Shuffle and MIL Head.}
To evaluate the individual contributions of Cross-Shuffle and the MIL Head, we conduct component ablations on GenImage. As shown in Table~\ref{tab:ablation1}, employing the MIL Head alone leads to a noticeable performance drop in overall accuracy, primarily due to misclassifications on fake images. In contrast, applying Cross-Shuffle alone substantially boosts detection accuracy, while combining both components yields the highest overall accuracy. These results demonstrate that the MIL Head relies on the diverse instance compositions provided by Cross-Shuffle to effectively capture and aggregate localized synthetic artifacts. Notably, full LTF achieves a substantial gain in fake-class accuracy over the baseline, providing strong empirical evidence for the principle of Learning to Falsify.

\noindent\textbf{Effect of Hierarchical Image Representation.}
To validate our hierarchical modeling, we compare the tile-level branch, patch-level branch, and their hierarchical fusion. As shown in Table~\ref{tab:ablation2}, hierarchical fusion achieves the best overall performance with an accuracy of 98.40\%, outperforming both individual branches and demonstrating their complementary nature for AIGI detection. Although the patch-level branch attains high accuracy on fake images, it suffers from a noticeably lower accuracy on real images, exhibiting a higher false-positive rate. By incorporating the tile-level representation, hierarchical fusion effectively restores performance on real images while maintaining strong fake-image detection, achieving a superior balance across both classes.

\begin{table}[t]
\centering
\caption{Ablation of the core LTF components on GenImage. All metrics are reported in percentage (\%).}
\label{tab:ablation1}
\small
\renewcommand{\arraystretch}{1.15}

\begin{tabular*}{0.96\columnwidth}
{@{\extracolsep{\fill}}lccccc@{}}
\toprule
\textbf{Model} &
\textbf{mAcc} &
\textbf{Real Acc} &
\textbf{Fake Acc} &
\textbf{AUC} &
\textbf{AP} \\
\midrule
Baseline
& 91.86
& 97.17
& 86.55
& 98.22
& 98.17 \\

+ MIL Head
& 87.93
& \textbf{99.89}
& 75.97
& 99.51
& 99.48 \\

+ Cross Shuffle
& \underline{96.41}
& \underline{98.77}
& \underline{94.06}
& \underline{99.63}
& \underline{99.62} \\

\addlinespace[1pt]
\textbf{Full LTF}
& \textbf{98.40}
& 97.68
& \textbf{99.09}
& \textbf{99.89}
& \textbf{99.86} \\
\bottomrule
\end{tabular*}
\end{table}
\begin{table}[t]
\centering
\caption{Ablation of hierarchical evidence modeling in LTF on GenImage. All metrics are reported in percentage (\%).}
\label{tab:ablation2}
\small
\renewcommand{\arraystretch}{1.15}

\begin{tabular*}{0.96\columnwidth}
{@{\extracolsep{\fill}}lccccc@{}}
\toprule
\textbf{Branch} &
\textbf{mAcc} &
\textbf{Real Acc} &
\textbf{Fake Acc} &
\textbf{AUC} &
\textbf{AP} \\
\midrule
Tile level
& 95.26
& 92.41
& 98.12
& 99.36
& \underline{99.34} \\

Patch level
& \underline{97.39}
& \underline{95.24}
& \textbf{99.55}
& \underline{99.88}
& \textbf{99.86} \\

\textbf{Hierarchy}
& \textbf{98.40}
& \textbf{97.68}
& \underline{99.09}
& \textbf{99.89}
& \textbf{99.86} \\
\bottomrule
\end{tabular*}
\end{table}

\subsection{Robustness Analysis}

\begin{figure}
    \centering
    \includegraphics[width=\linewidth]{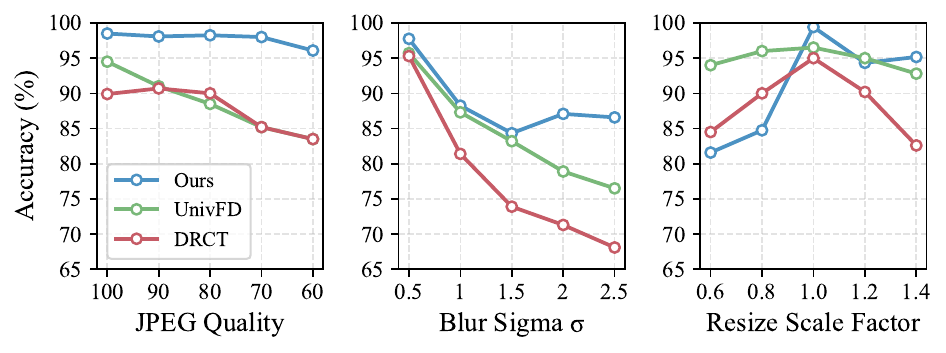}
    \caption{Robustness comparison under JPEG compression, Gaussian blur, and image rescaling. Our method achieves more stable performance overall.}
    \Description{Three line charts compare the accuracy of the proposed method, UnivFD, and DRCT under image degradation. In the JPEG chart, quality decreases from 100 to 60; the proposed method stays near 96 to 99 percent, while both baselines fall to roughly 84 percent at quality 60. In the Gaussian-blur chart, blur sigma increases from 0.5 to 2.5. All methods begin near 95 to 97 percent, but the proposed method remains around 83 to 88 percent after sigma 1.0, while UnivFD decreases to about 76 percent and DRCT to about 68 percent at sigma 2.5. In the rescaling chart, scale factors range from 0.6 to 1.4. The proposed method is weaker than UnivFD under downscaling at 0.6 and 0.8, peaks near 99 percent at the original scale of 1.0, and stays in the mid-90 percent range for enlargement. DRCT drops sharply at the largest scale, while UnivFD varies more moderately.}
    \label{fig:robustness}
\end{figure}

To evaluate robustness to common image degradations, we compare our method with UnivFD and DRCT on GenImage under three types of image transformations: JPEG compression, Gaussian blur, and image rescaling. As shown in Fig.~\ref{fig:robustness}, our method demonstrates stronger robustness than UnivFD and DRCT across these degradation settings. Under JPEG compression, it outperforms both baselines by approximately 13 percentage points at a quality factor of 60. Under Gaussian blur, our method consistently achieves the highest accuracy, outperforming UnivFD and DRCT by 10-18 percentage points under severe blur ($\sigma=2.5$). Although its performance is less competitive under downscaling, where fine-grained details are lost, our method achieves the highest accuracy among the compared methods when the scale factor exceeds 1.0. 

\begin{table}[t]
\centering
\caption{Efficiency comparison on HiRes-50K.}
\label{tab:complexity}
\setlength{\tabcolsep}{3.5pt}
\renewcommand{\arraystretch}{1.12}
\small
\begin{tabular}{lccccc}
\toprule
\textbf{Method} &
\makecell{\textbf{FLOPs}\\\textbf{(G)} $\downarrow$} &
\makecell{\textbf{Latency}\\\textbf{(ms)} $\downarrow$} &
\makecell{\textbf{Memory}\\\textbf{(MB)} $\downarrow$} &
\makecell{\textbf{Params}\\\textbf{(M)} $\downarrow$} &
\makecell{\textbf{Accuracy}\\\textbf{(\%)} $\uparrow$} \\
\midrule

\multicolumn{6}{l}{\textit{Single-patch methods}} \\
\cmidrule(lr){1-6}
UnivFD & 77.83  & 10.60 & 1699 & 427.6 & 61.99 \\
DRCT   & 77.83  & 10.60 & 1664 & 428.4 & 67.17 \\
AIDE   & 225.70 & 32.90 & 3425 & 897.8 & 56.46 \\

\addlinespace[2pt]
\multicolumn{6}{l}{\textit{Multi-patch methods}} \\
\cmidrule(lr){1-6}
SPAI & 658.00 & 233.00 & 4376 & 139.95 & 69.92 \\
HiDA & 821.16 & 64.26  & 1714 & 403.0  & 68.53 \\
\rowcolor{gray!12}
\textbf{LTF} & 779.54 & 57.17 & 1395 & 319.5 & 81.43 \\

\bottomrule
\end{tabular}
\end{table}
\subsection{Computational Efficiency}
We evaluate computational efficiency on the same machine under identical settings. For a fair comparison, all multi-patch methods are standardized to a fixed budget of 10 patches per image. As shown in Table~\ref{tab:complexity}, single-patch methods are faster but less accurate, whereas existing multi-patch approaches trade high computational overhead for accuracy gains. In contrast, LTF achieves a superior balance between computational efficiency and detection performance.

\section{Conclusion}
In this work, we introduced Learning to Falsify as a general principle for AI-generated image detection. Based on this principle, we developed LTF, a hierarchical MIL framework that learns discriminative representations across multiple levels. Cross-Shuffle is a key component of LTF, as it creates diverse instance compositions that enable the MIL Head to better capture localized artifacts. We also proposed FATAS, a plug-and-play method that enhances existing detectors through multi-patch inference and falsification-driven aggregation. Extensive experiments on GenImage, DRCT, Chameleon, and HiRes-50K demonstrate the strong cross-dataset generalization of our approach, particularly on high-resolution images.

Despite its favorable accuracy-efficiency trade-off, our approach remains more computationally demanding than single-patch methods.
Improving its efficiency is therefore an important direction for future research. Beyond image detection, we believe the Learning to Falsify principle can also be extended to AI-generated video detection, where we expect similar potential for improvement.

\begin{acks}
This work was supported by the National Natural Science Foundation of China (No. 62472385), NSFC (Joint Fund) Key Program (No. U25A20440), Pioneer and Leading Goose R\&D Program of Zhejiang (No. 2024C01110),  Fundamental Research Funds for the Provincial Universities of Zhejiang (No. FR2402ZD), Tongxiang Institute of Artificial General Intelligence (No. TAGI2-B-2024-0004), Zhejiang Provincial High-Level Talent Special Support Program (2022R52048), the Graduate Scientific Research and Innovation Fund of Zhejiang Gongshang University (No. ZDXM2025010), Zhejiang Provincial Science and Technology Innovation Program (New Young Talent Program) for College Students (No. 2025R408A008) and the Computing Power Resource of the Future Experiment Center of Zhejiang Gongshang University.
\end{acks}

\balance
\bibliographystyle{ACM-Reference-Format}
\bibliography{reference}

\end{document}